\documentstyle{article}

\begin{document}
\pagestyle{myheadings}
\markright{{\rm Coling 1994 -- Karlgren, Cutting:
           Recognizing Text Genres...} \hfill }

\title{\large \sc Recognizing Text Genres with Simple Metrics Using
Discriminant Analysis}

\author{\normalsize
\begin{tabular}{cc}
{\sc Jussi Karlgren } \mbox{\hspace{1cm}} & {\sc Douglass Cutting }    \\
\verb!jussi@sics.se!  \mbox{\hspace{1cm}} & \verb!cutting@apple.com!   \\
Swedish Institute of Computer Science
                      \mbox{\hspace{1cm}}  & Apple Computer            \\
Box 1263, S -- 164 28 {\sc Kista}, Stockholm, Sweden
                       \mbox{\hspace{1cm}}    & Cupertino, CA 95014, USA    \\
\end{tabular}
}

\maketitle

\begin{center}
\subsection*{Abstract}
\end{center} {\small
A simple method for categorizing texts into pre-de\-termined text genre
categories using the statistical standard technique of discriminant
analysis is demonstrated with application to the Brown corpus. Discriminant
analysis makes it possible use a large number of parameters that
may be specific for a certain corpus or information stream, and combine
them into a small number of functions, with the parameters weighted
on basis of how useful they are for discriminating text genres. An
application to information retrieval is discussed. }

\begin{center}
\subsection*{Text Types}
\end{center}
There are different types of text. Texts ``about'' the same thing may be in
differing genres, of different types, and of varying quality. Texts vary
along several parameters, all relevant for the general information
retrieval problem of matching reader needs and texts.  Given this variation,
in a text retrieval context the problems are (i) identifying genres,
and (ii) choosing criteria to cluster texts of the same genre, with
predictable precision and recall. This should not be confused with the
issue of identifying topics, and choosing criteria that discriminate one
topic from another. Although not orthogonal to genre-dependent
variation, the variation that relates directly to content and topic is
along other dimensions. Naturally, there is co-variance. Texts about
certain topics may only occur in certain genres, and texts in certain
genres may only treat certain topics; most topics do, however, occur in
several genres, which is what interests us here.

Douglas Biber has studied text variation along several parameters, and
found that texts can be considered to vary along five dimensions. In his
study, he clusters features according to covariance, to find underlying
dimensions (1989). We wish to find a method for identifying easily
computable parameters that rapidly classify previously unseen texts in
general classes and along a small set -- smaller than Biber's five -- of
dimensions, such that they can be explained in intuitively simple terms to
the user of an information retrieval application. Our aim is to take a set
of texts that {\em has} been selected by some sort of crude semantic analysis
such as is typically performed by an information retrieval system and
partition it {\em further} by genre or text type, and to display this variation
as simply as possible in one or two dimensions.

\begin{center}
\subsection*{Method}
\end{center}
We start by using features similar to those first investigated by Biber,
but we concentrate on those that are easy to compute assuming we have a
part of speech tagger (Cutting {\it et al}, 1992; Church, 1988), such as
such as third person pronoun occurrence rate as opposed to 'general hedges'
(Biber, 1989).  More and more of Biber's features will be available with
the advent of more proficient analysis programs, for instance if complete
surface syntactic parsing were performed before categorization (Voutilainen
\& Tapanainen, 1993).

We then use discriminant analysis, a technique from descriptive statistics.
Discriminant analysis takes a set of precategorized individuals and data on
their variation on a number of parameters, and works out a set {\em
discriminant functions} which distinguishes between the groups.  These
functions can then be used to predict the category memberships of new
individuals based on their parameter scores (Tatsuoka, 1971; Mustonen,
1965).

\begin{table}
{\small
\centering
\begin{tabular}{|l|l|l|}
\hline
Experiment 1             & Experiment 2      & Experiment 3 \\
                         &                   & {\small (Brown categories)} \\
\hline
I. Informative           & 1. Press          & A. Press: reportage  \\
\cline{3-3}
                         &                   & B. Press: editorial     \\
\cline{3-3}
	                 &                   & C. Press: reviews        \\ \cline{2-3}
	                 & 4. Misc           & D. Religion   \\  \cline{3-3}
	                 &                   & E. Skills and Hobbies    \\
\cline{3-3}
	                 &                   & F. Popular Lore          \\
\cline{3-3}
	                 &                   & G. Belles Lettres, etc.\\ \cline{2-3}
	                 & 2. Non-fiction    & H. Gov. doc. \& misc. \\ \cline{3-3}
	                 &                   & J. Learned               \\
\hline
II. Imaginative          & 3. Fiction        & K. General Fiction      \\
\cline{3-3}
	                 &                   & L. Mystery \\ \cline{3-3}
	                 &                   & M. Science Fiction  \\ \cline{3-3}
	                 &                   & N. Adv. \& Western  \\ \cline{3-3}
	                 &                   & P. Romance  \\ \cline{3-3}
	                 &                   & R. Humor \\
\hline
\end{tabular}
\caption{Categories in the Brown Corpus}\label{BrownCats}
}
\end{table}

\begin{table}
\centering
{\small
\begin{tabular}{|ll|}
\hline
Variable & Range  \\
\hline
Adverb count  			& 19 -- 157 		\\
Character count  		& 7601 -- 12143 	\\
Long word count ($>$ 6 chars) 	& 168 -- 838  		\\
Preposition count  		& 151 -- 433 		\\
Second person pronoun count  	& 0 -- 89 		\\
``Therefore'' count  		& 0 -- 11 		\\
Words per sentence average  	& 8.2 -- 53.2 		\\
Chars / sentence average        & 34.6 -- 266.3 	\\
First person pronoun count  	& 0 -- 156 		\\
``Me'' count  			& 0 -- 30 		\\
Present participle count	& 6 -- 101 		\\
Sentence count  		& 40 -- 236 		\\
Type / token ratio  		& 14.3 -- 53.0 		\\
``I'' count  			& 0 -- 120 		\\
Character per word average  	& 3.8 -- 5.8 		\\
``It'' count  			& 1 -- 53 		\\
Noun count  			& 243 -- 751 		\\
Present verb count 		& 0 -- 79 		\\
``That'' count  		& 1 -- 72 		\\
``Which'' count  		& 0 -- 40 		\\
\hline
\end{tabular}
}

\caption{Parameters for Discriminant Analysis}\label{Feats}
\end{table}

\begin{center}
\subsection*{Evaluation}
\end{center}
For data we used the Brown corpus of English text samples of uniform
length, categorized in several categories as seen in table~\ref{BrownCats}.
We ran discriminant analysis on the texts in the corpus using several
different features as seen in table~\ref{Feats}. We used the SPSS system
for statistical data analysis, which has as one of its features a complete
discriminant analysis (SPSS, 1990).  The discriminant function extracted
from the data by the analysis is a linear combination of the parameters. To
categorize a set into $N$ categories $N-1$ functions need to be determined.
However, if we are content with being able to plot all categories on a
two-dimensional plane, which probably is what we want to do, for ease of
exposition, we only use the two first and most significant functions.

\begin{center}
\subsubsection*{2 categories}
\end{center}
In the case of two categories, only one function is necessary for
determining the category of an item. The function classified 478 cases
correctly and misclassified 22, out of the 500 cases, as shown in
table~\ref{2Cats} and figure~\ref{Map2Cats}.

\begin{center}
\begin{table*}
\centering
\begin{tabular}{|l|l|r|}
\hline
Category  	 & Items	& Errors \\
\hline
I.  Informative 		& 374 &  16 (4 \%) \\
II. Imaginative 		& 126 &   6 (5 \%)    \\
\hline
Total				& 500 &  22 (4 \%) \\
\hline
\end{tabular}
\caption{Categorization in Two Categories}\label{2Cats}
\end{table*}

\begin{figure}
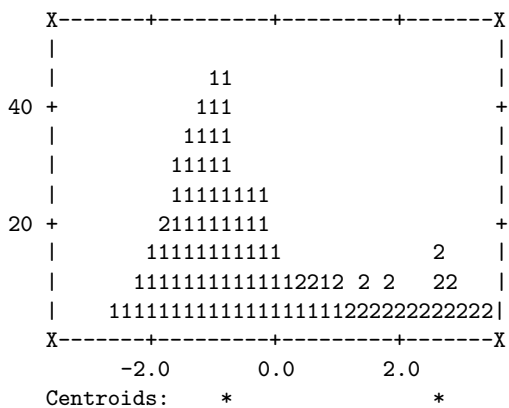

{\small \centering
\begin{verbatim}
     X-------+---------+---------+-------X
     |                                   |
     |            11                     |
  40 +           111                     +
     |          1111                     |
     |         11111                     |
     |         11111111                  |
  20 +        211111111                  +
     |       11111111111            2    |
     |      11111111111112212 2 2   22   |
     |    1111111111111111111222222222222|
     X-------+---------+---------+-------X
           -2.0       0.0       2.0
     Centroids:    *                *
\end{verbatim}
}
\caption{Distribution, 2 Categories}\label{Map2Cats}
\end{figure}
\end{center}

\begin{center}
\begin{figure}
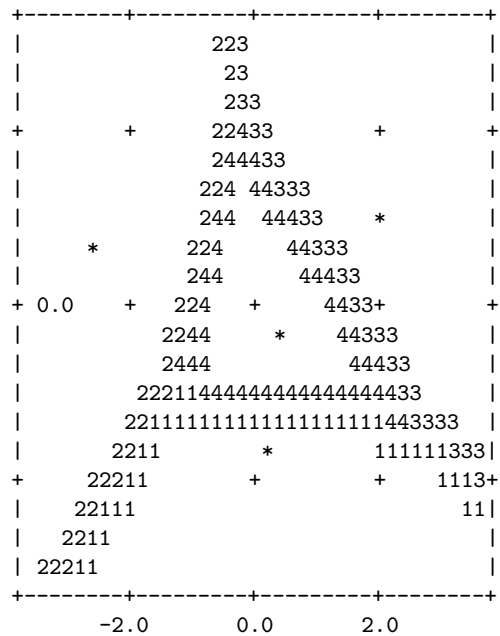

{\small \centering
\begin{verbatim}
   +--------+---------+---------+--------+
   |               223                   |
   |                23                   |
   |                233                  |
   +        +      22433        +        +
   |               244433                |
   |              224 44333              |
   |              244  44433    *        |
   |     *       224     44333           |
   |             244      44433          |
   + 0.0    +   224   +     4433+        +
   |           2244     *    44333       |
   |           2444           44433      |
   |         22211444444444444444433     |
   |        221111111111111111111443333  |
   |       2211        *        111111333|
   +     22211        +         +    1113+
   |    22111                          11|
   |   2211                              |
   | 22211                               |
   +--------+---------+---------+--------+
          -2.0       0.0       2.0
\end{verbatim}
}
\caption{Distribution, 4 Categories}\label{Map4Cats}
\end{figure}
\end{center}

\begin{center}
\subsubsection*{4 categories}
\end{center}
Using the three functions extracted, 366 cases were correctly classified,
and 134 cases were misclassified, out of the 500 cases, as can be seen in
table~\ref{4Cats} and figure~\ref{Map4Cats}. ``Miscellaneous'', the most
problematic category, is a loose grouping of different informative
texts.  The single most problematic subsubset of texts is a subset of
eighteen non-fiction texts labeled ``learned/humanities''. Sixteen of them
were misclassified, thirteen as ``miscellaneous''.

\begin{table}
\small \centering
\begin{tabular}{|l|l|r|}
\hline
Category        & Items	& Errors  \\
\hline
1. Press 	&  88 &  15 (17 \%)   \\
2. Non-fiction 	& 110 &  28 (25 \%)     \\
3. Fiction   	& 126 &   6 (5 \%)     \\
4. Misc.        & 176 &  68 (47 \%)     \\
\hline
Total		& 500 & 134 (27 \%)  \\
\hline
\end{tabular}
\caption{Categorization in Four Categories}\label{4Cats}
\end{table}

\begin{center}
\subsubsection*{15 (or 10) categories}
\end{center}
Using the fourteen functions extracted, 258 cases were correctly classified
and 242 cases misclassified out of the 500 cases, as shown in
table~\ref{15Cats}. Trying to distinguish between the different types of
fiction is expensive in terms of errors. If the fiction subcategories were
collapsed there only would be ten categories, and the error rate for the
categorization would improve as shown in the ``revised total'' record of
the table. The ``learned/humanities'' subcategory is, as before, problematic:
only two of the eighteen items were correctly classified. The others
were most often misclassified as ``Religion'' or ``Belles Lettres''.

\begin{center}
\subsection*{Validation of the Technique}
\end{center}
It is important to note that this experiment does not claim to show {\em
how} genres in fact differ. What we show is that this sort of technique
{\em can be} used to determine which parameters to use, given a set of
them. We did not use a test set disjoint from the training set, and we do
not claim that the functions we had the method extract from the data are
useful in themselves. We discuss how well this method categorizes a set
text, given a set of categories, and given a set of parameters.

The error rates climb steeply with the number of categories tested for in
the corpus we used. This may have to do with how the categories are chosen
and defined. For instance, distinguishing between different types of
fiction by formal or stylistic criteria of this kind may just be something
we should not attempt: the fiction types are naturally defined in terms of
their content, after all.

The statistical technique of {\em factor analysis} can be used to discover
categories, like Biber has done. The problem with using automatically
derived categories is that even if they are in a sense real, meaning that
they are supported by data, they may be difficult to explain for the
unenthusiastic layman if the aim is to use the technique in retrieval
tools.

Other criteria that should be studied are second and higher order
statistics on the respective parameters. Certain parameters probably {\em
vary more} in certain text types than others, and they may have a {\em
skewed distribution} as well. This is not difficult to determine, although
the standard methods do not support automatic determination of standard
deviation or skewness as discrimination criteria. Together with the
investigation of several hitherto untried parameters, this is a next step.

\begin{center}
\subsection*{Readability Indexing}
\end{center}
Not unrelated to the study of genre is the study of {\em readability} which
aims to categorize texts according to their suitability for assumed sets of
assumed readers. There is a wealth of formul{\ae} to compute readability.
Most commonly they combine easily computed text measures, typically average
or sampled average sentence length combined with similarly computed word
length, or incidence of words not on a specified ``easy word list'' (Chall,
1948; Klare, 1963). In spite of Chall's warnings about injudicious
application to writing tasks, readability measurement has naively come to
be used as a prescriptive metric of good writing as a tool for writers, and
has thus come into some disrepute among text researchers. Our small study
confirms the basic findings of the early readability studies: the most
important factors of the ones we tested are word length, sentence length,
and different derivatives of these two parameters. As long as readability
indexing schemes are used in descriptive applications they work well to
discriminate between text types.

\begin{center}
\subsection*{Application}
\end{center}
The technique shows practical promise. The territorial maps shown in
figures~\ref{Map2Cats}, \ref{Map4Cats}, and \ref{Map15Cats} are intuitively
useful tools for displaying what type a particular text is, compared with
other existing texts. The technique demonstrated above has an obvious
application in information retrieval, for picking out interesting texts, if
content based methods select a too large set for easy manipulation and
browsing (Cutting {\it et al}, 1992).

In any specific application area it will be unlikely that the text database
to be accessed will be completely free form. The texts under consideration
will probably be specific in some way. General text types may be useful,
but quite probably there will be a domain- or field-specific text typology.
In an envisioned application, a user will employ a cascade of filters
starting with filtering by topic, and continuing with filters by genre or
text type, and ending by filters for text quality, or other tentative
finer-grained qualifications.

\begin{center}
\subsection*{The IntFilter Project}
\end{center}
The IntFilter Project at the departments of Computer and Systems Sciences,
Computational Linguistics, and Psychology at Stockholm University is at
present studying texts on the USENET News conferencing system. The project
at present studies texts which appear on several different types of USENET
News conferences, and investigates how well the classification criteria and
categories that experienced USENET News users report using (IntFilter,
1993) can be used by a newsreader system. To do this the project applies
the method described here. The project uses categories such as ``query'',
``comment'', ``announcement'', ``FAQ'', and so forth, categorizing them
using parameters such as different types of length measures, form word content,
quote level, percentage quoted text and other USENET News specific parameters.

\begin{center}
\begin{table*}
\centering
\begin{tabular}{|l|l|l|r|l|}
\hline
\small
Category  	 & Items	& Errors  & Miss \\
\hline
A. Press: reportage 			& 44 & 11 (25 \%) & F	\\
B. Press: editorial 			& 27 &  8 (30 \%) & A	\\
C. Press: reviews   			& 17 &  4 (24 \%) & B      \\
D. Religion  				& 17 &  8 (47 \%) & G	\\
E. Skills and Hobbies  			& 36 & 17 (47 \%) & J	\\
F. Popular Lore        			& 48 & 32 (67 \%) & G,E	\\
G. Belles Lettres, Biographies etc. 	& 75 & 49 (65 \%) & D,B,A	\\
H. Government documents \& misc. 	& 30 &  9 (30 \%) & J	\\
J. Learned               		& 80 & 32 (40 \%) & H,D,G,F \\
K. General Fiction      		& 29 & 16 (55 \%) & fiction \\
L. Mystery 				& 24 & 12 (50 \%) & -"- \\
M. Science Fiction  			&  6 &  1 (17 \%) & -"- \\
N. Adventure and Western  		& 29 & 18 (62 \%) & -"-  \\
P. Romance   				& 29 & 22 (76 \%) & -"-  \\
R. Humor  				&  9 &  3 (33 \%) & -"-  \\
\hline
Total					& 500 & 242 (48 \%) & \\
\hline
Fiction {\small (From previous table)} 	& 126 &   6 (5 \%) &    \\
Revised total 				& 500 & 178 (35 \%) & \\
\hline
\end{tabular}
\caption{Categorization in 15 Categories}\label{15Cats}
\end{table*}

\begin{figure}
{\small \centering
\begin{verbatim}
  +---------+---------+---------+---------+---------+---------+
  |        -4        -2    LJJ  0         2         JHH       |
  |                        LLJ                      JJH       |
  |                       LLPJJ                      JH       |
  +         +         + LLLPKFJJ+         +         +JHH      +
  |                   LLLPKKKFFJJJ                   JJH      |
  |                 LLLPKKKKFFFFFJJJ     *  *         JHH     |
  |             * LLLPKKK  KF   FFFJJ                 JJH     |
  |            L**LNPKK    KF     FFJJJ                JH     |
  |         LLLLNNNKKK*   KKF      *FFJJJ              JHH    |
  +      LLLLNNNNKKK* +   KFF   +   *FFFJJ+         +  JJH    +
  |   LLLLNNNNNKKK        KF     *    FFFJJJJ           JHH   |
  |LLLLNNNNNNNKK         KKF       *FFFGGGGGJJJ         JJH   |
  |LNNNN NNNKKK        KK*RFFFFFFFFFFGGGG  GGGJJJJ       JH   |
  |NN   NNKKK         KKRRRBBBBBBB*BBBBBGGGGGGGGGJJJJ    JHH  |
  |   NNNKK         KKKRR RB       *   BBBBBBGGGGGGGJJJJ JJH  |
  + NNNKKK  +     KKKRRR RRB    +         + BBAAAAAAAAAJJJJHH +
  |NNKKK        KKKRRR   RBB      *         BBA       AAAAJJHH|
  |NKK         KKRRR     RB                BBAA          AAAAH|
  |KK        KKKRR      RRB                BAA              AA|
  |        KKKRRR       RBB               BBA                 |
  |      KKKRRR        RRB           BBBBBBAAAAAA             |
  +     KKRRR         +RBBBBBBBBBBBBBBCCCCCCCCCCAAAAAAAAAAAAAA+
  |   KKKRR           RRBBBCCCCCCCCCCCC        CCCCCCCCCCCCCCC|
  | KKKRRR           RRCCCCC                                  |
  |KKRRR            RRCC                                      |
  +---------+---------+---------+---------+---------+---------+
\end{verbatim}
}
\caption{Distribution, 15 Categories
          -- * Indicates a group centroid.}\label{Map15Cats}
\end{figure}
\end{center}

\begin{center}
\subsection*{Acknowledgements}
\end{center}
Thanks to Hans Karlgren, Gunnel K\"allgren, Geoff Nunberg, Jan Pedersen,
and the Coling referees, who all have contributed with suggestions and
methodological discussions.
\newpage
\begin{center}
\subsection*{References}
\end{center}
{\small
\begin{description}
\item {\bf Douglas Biber} 1989.  ``A typology of English texts'',
                {\it Linguistics, 27}:3-43.

\item {\bf Jeanne S. Chall} 1948. {\it Readability}, Ohio State Univ.

\item {\bf Kenneth Church} 1988.
	``A Stochastic Parts of Speech and Noun Phrase Parser for
          Unrestricted Text'',
	{\it Procs.~2nd ANLP\/}, Austin.

\item {\bf Douglass Cutting, Julian Kupiec, Jan  Pedersen, and Penelope Sibun}
1992.
         \hfill
	``A Practical Part-of-Speech Tagger'',
	{\it Procs.~3rd ANLP\/}, Trento.

\item {\bf Douglass Cutting, D. Karger, Jan  Pedersen, and John Tukey} 1992.
``Scatter/Gather: A Cluster-based Approach to Browsing Large Document
Collections'' {\it Procs. SIGIR'92}.

\item {\bf IntFilter} 1993. \\ {\em Working Papers of the IntFilter Project},
                 available by gopher from
                 \verb!dsv.su.se:/pub/IntFilter!.

\item {\bf George R. Klare} 1963. {\em The Measurement of Read\-abi\-li\-ty},
                Iowa Univ press.

\item {\bf W. N. Francis and F. Ku\v{c}era} 1982.
 {\it Frequency Analysis of English Usage}, Houghton Mifflin.

\item {\bf Seppo Mustonen} 1965.
               ``Multiple Discriminant Analysis in Linguistic Problems'',
               {\it Statistical Methods in Linguistics, 4}:37-44.

\item {\bf M. M. Tatsuoka} 1971.  {\it Multivariate Analysis},
                   New York:John Wiley \& Sons.
\item {\bf Atro Voutilainen and Pasi Ta\-pa\-nai\-nen} 1993.
	``Ambiguity resolution in a reductionistic par\-ser'',
	{\it Procs.~6th European ACL\/}, Utrecht.

\item {\bf SPSS} 1990.  {\it The SPSS Reference Guide}, Chicago: SPSS Inc.

\end{description}
}
\end{document}